\documentclass[10pt,twocolumn,letterpaper]{article}

\usepackage{wacv}
\usepackage{times} %% I checked WACV template has times
\usepackage{epsfig}
\usepackage{graphicx}
\usepackage{amsmath}
\usepackage{amssymb}
\usepackage{booktabs}
\usepackage{multirow}
\usepackage{comment}
\usepackage{bbm}
\usepackage{authblk}
\usepackage[inline]{enumitem}
\usepackage{algorithm,algpseudocode}
\usepackage{diagbox,makecell}
\usepackage[table,x11names]{xcolor}
\newcommand{\myfirstpara}[1]{\noindent {\bf #1:}}
\newcommand{\mypara}[1]{\vspace{0.5em} \myfirstpara{#1}}
\newcommand*\rot{\rotatebox{90}}

\def\mada{\texttt{MADA}\xspace}
\def\labor{\texttt{LabOR}\xspace}
\def\cdal{\texttt{CDAL}\xspace}
\def\iou{\texttt{IoU}\xspace}
\def\miou{\texttt{mIoU}\xspace}
\def\cityscapes{\texttt{Cityscapes}\xspace}
\def\synthians{\texttt{Synthia}}
\def\synthia{\synthians\xspace}
\def\gtans{\texttt{GTA5}}
\def\gta{\gtans\xspace}
\def\sota{\texttt{SOTA}\xspace}

\def\wacvPaperID{506} % Enter the WACV Paper ID here

\wacvalgorithmstrack   % Uncomment this line if you are submitting to the Algorithms Track.
\wacvfinalcopy % *** Uncomment this line for the final submission

\ifwacvfinal
\usepackage[breaklinks=true,bookmarks=false]{hyperref}
\else
\usepackage[pagebackref=true,breaklinks=true,colorlinks,bookmarks=false]{hyperref}
\fi

\usepackage[capitalise]{cleveref}

\hypersetup{
    colorlinks=true,
    linkcolor=red,
    filecolor=magenta,      
    urlcolor=red,
    pdftitle={Overleaf Example},
    pdfpagemode=FullScreen,
    }
\begin{document}

%%%%%%%%% TITLE
\title{Reducing Annotation Effort by Identifying and Labeling Contextually Diverse Classes for Semantic Segmentation Under Domain Shift}
\author[1]{Sharat Agarwal}
\author[1]{Saket Anand}
\author[2]{Chetan Arora}

\affil[1]{IIIT Delhi, India {\tt\small \{sharata,anands\}@iiitd.ac.in}}
\affil[2]{Indian Institute of Technology Delhi, India {\tt\small chetan@cse.iitd.ac.in}}

\maketitle
\thispagestyle{empty}
\begin{abstract}
In Active Domain Adaptation (ADA), one uses Active Learning (AL) to select a subset of images from the target domain, which are then annotated and used for supervised domain adaptation (DA). Given the large performance gap between supervised and unsupervised DA techniques, ADA allows for an excellent trade-off between annotation cost and performance.  Prior art makes use of measures of uncertainty or disagreement of models to identify `regions' to be annotated by the human oracle. However, these regions frequently comprise of pixels at object boundaries which are hard and tedious to annotate. Hence, even if the fraction of image pixels annotated reduces, the overall annotation time and the resulting cost still remain high. In this work, we propose an ADA strategy, which given a frame, identifies a set of classes that are hardest for the model to predict accurately, thereby recommending semantically meaningful regions to be annotated in a selected frame. We show that these set of `hard' classes are context-dependent and typically vary across frames, and when annotated help the model generalize better. We propose two ADA techniques: the \texttt{Anchor-based} and \texttt{Augmentation-based} approaches to select complementary and diverse regions in the context of the current training set. Our approach achieves $66.6$ \miou on \gta$\rightarrow$\cityscapes dataset with an annotation budget of $4.7\%$ in comparison to $64.9$ \miou by \mada \cite{mada} using $5\%$ of annotations. Our technique can also be used as a decorator for any existing frame-based AL technique, e.g., we report $1.5\%$ performance improvement for \cdal~\cite{cdal} on \cityscapes using our approach.

\end{abstract}

\section{Introduction}
\label{sec:intro}

\begin{figure}[t!]
	\centering
	\includegraphics[width=\linewidth]{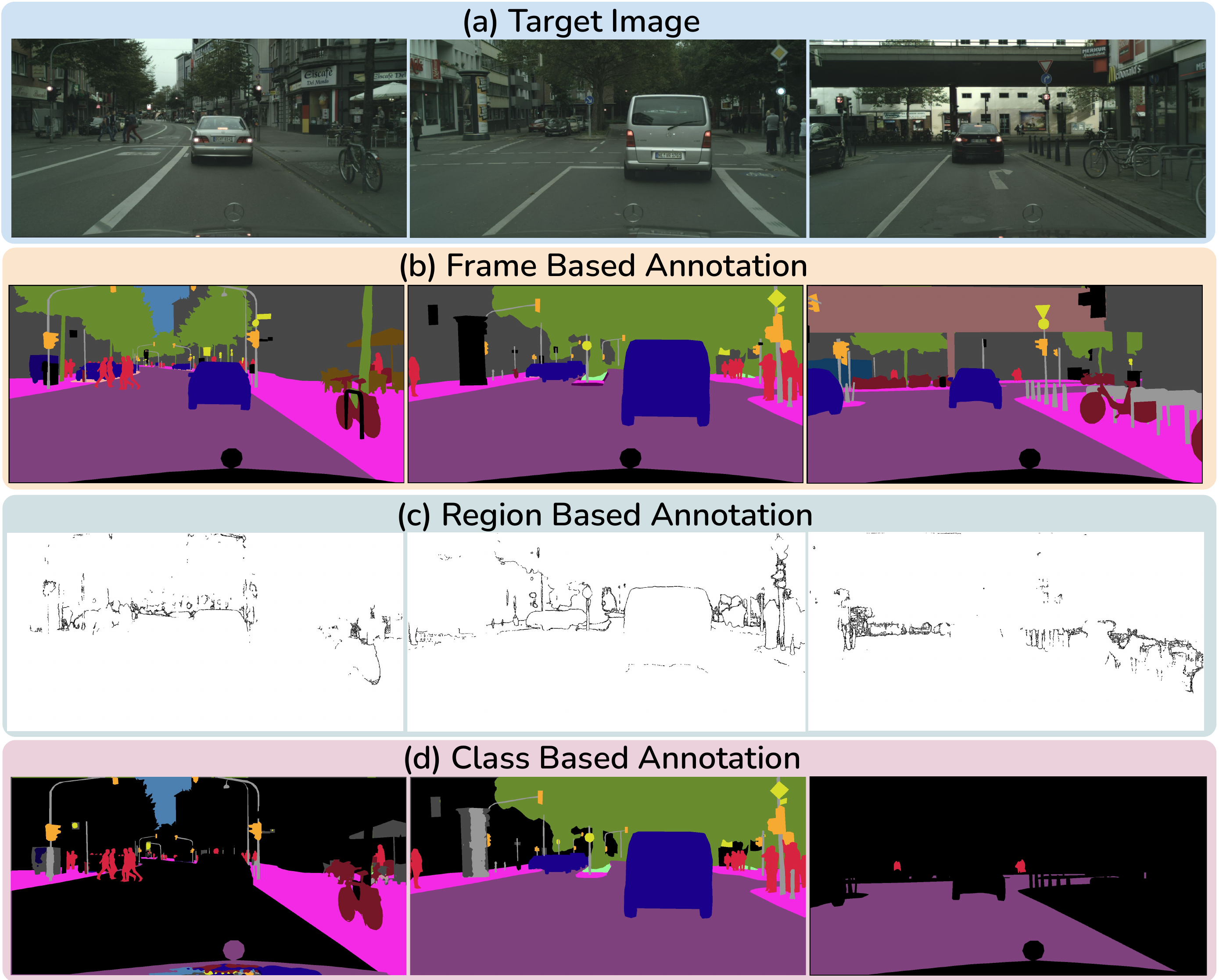}
	\caption{Different annotation strategies for semantic segmentation in active learning. (a) \textbf{Target Image}. (b) \textbf{Frame-based} methods annotate the entire selected frame, which leads to annotating redundant pixels wasting annotation budget. (c) \textbf{Region-Based} methods select arbitrary regions based on uncertainty in each image across the unlabeled pool, which are often tedious and challenging for a human to annotate. We propose a novel and intuitive approach, (d) \textbf{Class-based} annotation, where contextually relevant classes that are complementary for model training are identified in each actively selected frame, thus reducing the annotation effort and simultaneously increasing the model performance.}
	\label{fig:Teaser}
\end{figure}

One of the major stumbling blocks for development of robust semantic segmentation techniques is the associated cost for obtaining annotated samples from a variety of target domains. For instance, it takes 90 minutes to annotate a single \cityscapes \cite{cityscapes} image. One of the ways to mitigate data annotation effort/cost is by synthetically generating data with dense labels using 3D simulation platforms and game engines \cite{gta5,synthia}. However, Deep Neural Networks (DNNs) trained entirely on the synthetic data fail to generalize well in the real-world setting due to the domain shift. To address this issue several unsupervised  \cite{adaptnet, advent, luo2019taking, pycda, iast, wda}, semi-supervised \cite{chen2021semi, wang2020alleviating, saito2019semi} and self-training \cite{mei2020instance, zhang2021prototypical, zou2018unsupervised} based DA techniques were proposed. However, despite the immense effort, the performance still lags far behind the fully supervised models. 

To maximize model performance with minimal labeling effort, recently, ADA techniques \cite{mada, labor} have  been proposed where the most \emph{informative} samples from the unlabeled target domain are selected for the annotation. \mada \cite{mada}, a frame selection approach, labels target samples that are most complementary to source domain anchors. \labor \cite{labor} is a pixel-based approach which obtains labels for those uncertain pixels in an image where two classifiers disagree on their predictions. Both the approaches have shown significant performance improvement, but are highly inefficient in terms of annotation cost. On one hand, \mada wastes annotation budget by labeling redundant pixels in a selected frame (c.f. \cref{fig:Teaser}(b)). On the other hand, \labor selects sparse pixels belonging to different classes (c.f. \cref{fig:Teaser}(c)), which are tedious and time-consuming to annotate by a human annotator. By choosing to work at a pixel level, where diversity is hard to compute, \labor fails to consider annotated regions across the frames. 

We argue that to gain cost efficiency, it is critical to select semantically meaningful target domain  regions which are hard/novel for a model. Choosing the semantic regions rather than individual pixels maintains the simplicity of annotation task for a human oracle, and at the same time allows an automated algorithm to use dataset-wide criterion such as diversity, and novelty. E.g., if a model has only seen straight roads for the {\tt road} label, it will likely struggle in images containing road with turns, thus making {\tt road} as a \emph{hard} label in those frames. It is important to point out the contextual nature of the \emph{hard} classes, which may be quite different from the difficulty arising from the imbalanced classes. Consequently, we take an intuitive approach and define semantically meaningful regions as instances of the hard classes in a frame for the annotation. 

\mypara{Contributions}
\begin{enumerate*}[label=\textbf{(\arabic*)}]
	\item We introduce two ADA techniques for selecting hard classes in the frames selected by any contemporary frame selection strategy such as \cdal \cite{cdal}. Our \texttt{Anchor-based}  approach (c.f. \cref{sec:anchor-based}) selects class instances novel with respect to class-wise anchors chosen from the target dataset. This helps strengthen the per class representation in the feature space. Our \texttt{Augmentation-based} (c.f. \cref{sec:aug-based}) approach follows the self-supervised uncertainty estimation, and chooses class instances based on the disagreement in the prediction probabilities for the corresponding weakly and strongly corrupted samples.
	\item To specifically understand our contribution of choosing semantic regions instead of pixels or frames in ADA, we ablate using an hypothetical \texttt{\iou-based} (\cref{sec:iou-based}) selection approach. Here, we select low confident classes in a frame based on the difference in their \iou values from the current model to a hypothetical fully supervised model. We show that using this approach one can achieve performance equal to a fully supervised model using only $\sim7\%$ of the annotated data. This validates the significance of choosing semantic regions. %We give detailed analysis in the supplementary. 
	\item We compare with state-of-the-art (\sota) UDA and ADA techniques, reducing the error margin (difference in \miou from the  fully supervised model) from $4.7$ to $3$ in {\gtans$\rightarrow$\cityscapes} and from $5.7$ to $3.2$ in {\synthians$\rightarrow$\cityscapes} at an annotation budget of merely 5\%.
\end{enumerate*}
Complete source-code and pre-trained models for this work will be publicly released post-acceptance.

\begin{figure*}[ht]
    \centering
    \includegraphics[width=\linewidth, height=8cm]{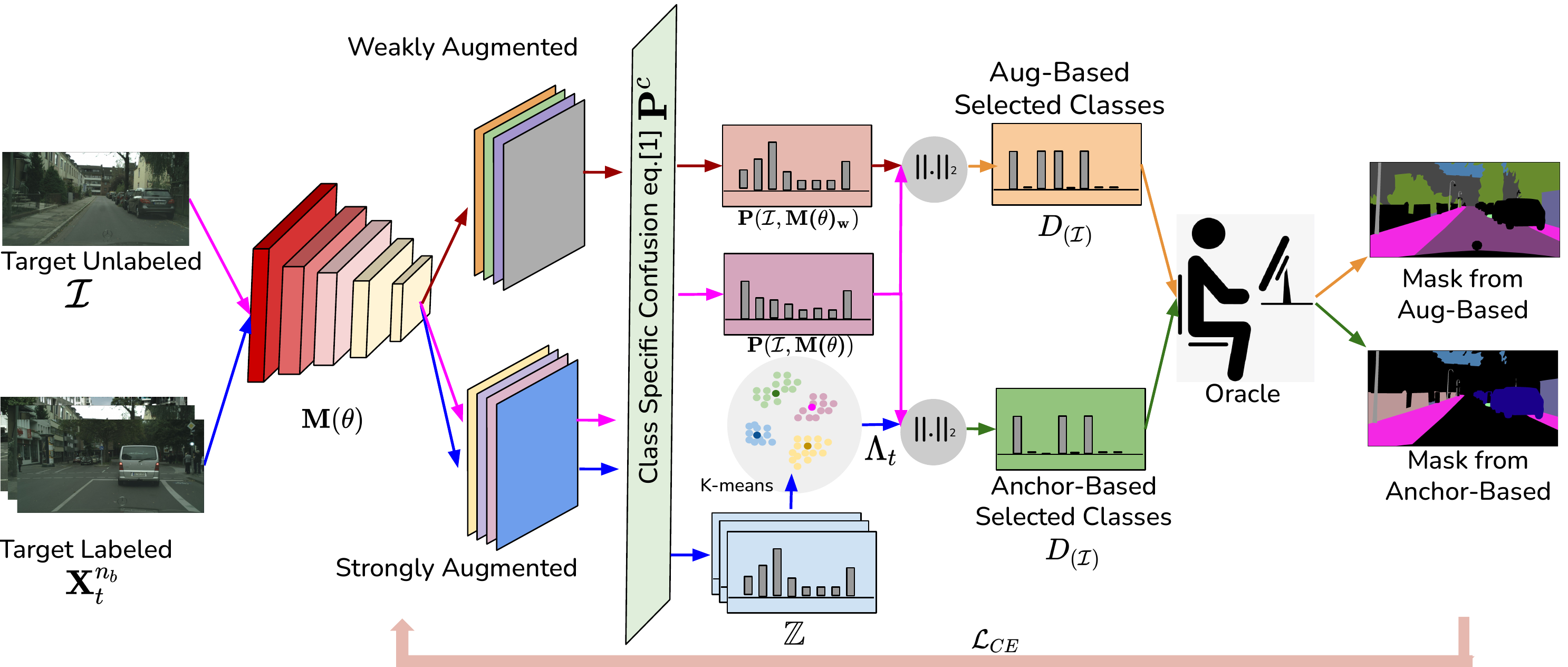}
    \caption{Overview of our proposed approach. Given small set of labeled target images $\mathbf{X}_{t}^{n_{b}}$ and an unlabeled image $\mathcal{I}$, \texttt{Anchor-based} and \texttt{Aug-based}  select classes $D_{\mathcal{I}}$ to be labeled. In \texttt{Anchor-based} we select classes complementary to the target anchors exploiting labeled target class distribution. Whereas, in \texttt{Aug-based} we measure the dissimilarity in the class confusions at inference time from Strong and Weakly augmented model. (Best visible in color)}
    \label{fig:Aug_method}
\end{figure*}

\section{Related Work}

%This section briefly discusses the literature on different types of domain adaptation approaches with no or few annotations and highlights the difference with our proposed approach. 

\myfirstpara{Domain Adaptation}
Unsupervised Domain Adaptation (UDA) addresses the problem of domain shift between the labeled source and unlabeled target domains and has been extensively explored for image classification \cite{ganin2015unsupervised}, object detection \cite{chen2018domain,tang2016large} and semantic segmentation \cite{adaptnet, advent, iast}. UDA is majorly categorized into two groups, based on (a) maximum mean discrepancy MMD \cite{long2015learning, long2017deep, tzeng2014deep} or (b) using adversarial learning \cite{adaptnet,advent,hong2018conditional}. Adversarial learning based approaches are more popular, and have been used to align source and target distributions at image~\cite{cycada,li2018semantic,toldo2020unsupervised}, feature~\cite{adaptnet,advent}, and output \cite{tsai2019domain,pan2020unsupervised} stages. 
Despite extensive interest, there is still a significant performance gap between supervised learning and UDA-based approaches \cite{mada}. To reduce the performance gap, semi-supervised learning (SSL) \cite{chen2021semi,wang2020alleviating,saito2019semi} based DA approaches have been proposed which utilize a small portion of randomly selected labeled target data for training. The implicit assumption is that the randomly selected set maintains the relationship between labeled and unlabeled data distribution \cite{lu2009fundamental}.

\mypara{Active Learning}
Instead of labeling randomly selected samples, AL algorithms choose the most valuable samples to be labeled by a human annotator \cite{settles2009active}. Since, annotating is far more expensive than collecting the data, several AL strategies have been proposed~\cite{ren2021survey}, based on ideas like membership query~\cite{king2004functional}, stream-based sampling \cite{krishnamurthy2002algorithms}, and pool-based sampling \cite{coreset,cdal}. Problems of interest include image classification~\cite{coreset,vaal}, object detection~\cite{cdal} and semantic segmentation~\cite{vaal,cdal}. Despite the enormous effort required in annotation for semantic segmentation, there has been a limited amount of work in this domain. 

\mypara{Active Domain Adaptation}
Active Domain adaptation (ADA) techniques adopt active learning for the task of domain adaptation, where most valuable samples from the unlabeled target domain are labeled. Recently, ADA techniques have been proposed for image classification \cite{prabhu2021active,rangwani2021s3vaada}. Our focus in this paper is on semantic segmentation, where the \sota techniques include \mada \cite{mada}, and \labor \cite{labor}. Whereas, \mada \cite{mada} annotates target domain frames most complementary to the anchors from the source domain, \labor \cite{labor} annotates most uncertain regions based on the classifier disagreement in each image. As highlighted in \cref{sec:intro} both the existing ADA approaches are highly inefficient in terms of annotation cost. %Thus we propose a simple but intuitive approach that efficiently prepares annotated data by selecting semantically meaningful classes in a selected frame. Our \texttt{Anchor-based} and \texttt{Aug-based} approaches helps in reducing the overall annotation budget and also increasing the overall model performance. 

\section{Methodology}

In this section we firstly discuss the preliminaries and the problem setup. Then, we present the proposed class selection approaches, \texttt{Aug-Based} (\ref{sec:aug-based}), \texttt{Anchor-Based} (\ref{sec:anchor-based}), the ground truth based skyline \texttt{IoU-Based} (\ref{sec:iou-based}), and finally our training objective (\ref{sec:training}). \cref{fig:Aug_method} shows the overview of our proposed approach.

\subsection{Problem Overview and Background}

In UDA, given the source dataset $\mathcal{X}_{s} = \{x_{s}\}^{n_{s}}$ with pixel-level labels $\mathcal{Y}_{s} = \{y_{s}\}^{n_{s}}$, the goal is to learn a segmentation model $\mathbf{M}(\theta)$ which can correctly predict pixel-level labels for the target domain samples, $\mathcal{X}_{t} = \{x_{t}\}^{n_{t}}$ without using $\mathcal{Y}_{t} = \{y_{t}\}^{n_{t}}$, where $\mathcal{Y}_{s}$ and $\mathcal{Y}_{t}$ share the same label space of $C$ classes and $n_s$ and $n_t$ are the number of images from the source and target domains. In Active Domain Adaptation (ADA) the task is to select a set of $n_b$ images in each iteration, $\mathbf{S}_{t}^{n_{b}} \subset \mathcal{X}_{t}^{n_{t}}$ with $n_{b} \ll n_{t}$ as the annotation budget, to be labeled by a human oracle such that $\mathbf{M}(\theta)$ achieves good performance on the target domain with only a few annotations.

Usually, traditional Active Learning approaches  either annotate the entire frame or annotate regions within a frame based on measures like uncertainty. Such measures are based on the model's performance, and may lead to semantically inconsistent regions that straddle class or object boundaries and are therefore harder to annotate manually. Contrary to this approach, we propose to select semantically meaningful regions as instances of certain classes in the selected frames to be annotated. Thus our \texttt{Anchor-based} and \texttt{Aug-based} identifies classes in the frames to be labeled. For our experiments we have used \cdal \cite{cdal}, but we further analyze the effectiveness of class selection in other frame selection techniques.

Let the frame selection function be $\Delta$, such that $\mathbf{S}_{t}^{n_{b}} = \Delta(\mathcal{X}_{t}^{n_{t}},n_{b})$, where $n_{b}$ is the annotation budget for each active learning iteration. The labeled pool is initialized as the set $\mathbf{X}_{t}[0]=\phi $ and in the $k^{th}$ iteration, is updated as $\mathbf{X}_{t}[k]=\{\mathbf{X}_{t}[k-1]\cup \mathbf{S}_{t}^{n_{b}}\}$. %, and the subset $\mathbf{S}_{t}^{n_{b}} $ is removed from the unlabeled set $\mathcal{X}_{t}^{n_{t}}$. 
In the first iteration, we initialize the model, $\mathbf{M}(\theta)$ with the warm-up weights from \cite{adaptnet}, and fine-tune it using the fully annotated frames from $\mathbf{X}_{t}[1]$. For every subsequent AL iteration, i.e., $k=2,3,\ldots$, $\Delta$ selects a fresh subset $\mathbf{S}_{t}^{n_{b}}$ from \{$\mathcal{X}_{t}^{n_{t}} \setminus \mathbf{X}_{t}[k]$\}, where $\setminus$ denotes the set difference operation. Our proposed class selection methods aim to select the most diverse and informative classes in each image $\mathcal{I} \in \mathbf{S}_{t}^{n_{b}}$, given the model trained on the most recent labeled set $\mathbf{X}_{t}[k] $. 

Given $\mathbf{M}(\theta)$ and image $\mathcal{I}$, we extract class specific confusion $\mathbf{P}^{c}$ for class $c \in C$ as proposed by \cite{cdal}. 
\begin{align}
\mathbf{P}^{c}({\mathcal{I}},\mathbf{M}(\theta)) = \frac{1}{|N_{c}|}\sum_{i\in N_{c}} \left [{\frac{w_{i} \times  \mathbf{p}_i[\hat{y}|\mathcal{I};\mathbf{M}(\theta)]}{\sum_{i\in N_c}{w_i}}}\right ]
\label{eq:1}
\end{align}
where $ N_c $ is the set of pixels that have been classified as class $ c $ and,
\begin{align}
    w_i = -\sum_{c \in C}\mathbf{p}_i[\hat{y}|\mathcal{I};\mathbf{M}(\theta)] \log_2 \mathbf{p}_i[\hat{y}|\mathcal{I};\mathbf{M}(\theta)]
\label{eq:2}
\end{align}
is the Shannon's entropy for the $ i^{th} $ pixel, and $\mathbf{p}_{i}$ is the softmax probability vector as predicted by the model $\mathbf{M}(\theta)$, and  $\widehat{y}$ is the random variable corresponding to the predicted class for a pixel $i$ in image $\mathcal{I}$. The class-specific confusion vector $\mathbf{P}^c({\mathcal{I}},\mathbf{M}(\theta))$ was introduced in \cite{cdal} as an entropy weighted mixture of softmax probabilities of the pixels predicted as class $c$. It can be interpreted as a weighted average of \emph{soft pseudo-labels}, where more uncertain pixels are assigned larger weights and highly confident pixels carry minimal weights. This weighted averaging of softmax probabilities results in a probability mass function that amplifies the probabilities of classes competing with class $c$, thus effectively capturing class confusion. Now we discuss our \texttt{Anchor-based} and \texttt{Aug-based} techniques, followed by \texttt{IoU-based} and training objective.

\subsection{{\tt Anchor-Based} Class Selection}
\label{sec:anchor-based}
Based on the observation that in the feature vectors corresponding to pixels of the same class belong to same cluster \cite{cag_uda}, we compute class representative \emph{anchors} at each AL iteration using the labeled target data annotated thus far. These class anchors capture the class-specific confusion in the model predictions, which in turn helps in selecting the most informative classes in the selected pool $\mathbf{S}_t^{n_b} $ which when included in the labeled set helps the model improve the overall class representation.

For computing the class representative anchors in the $k^{th}$ AL iteration, we construct a set of feature vectors $\mathbf{Z}^c$ using eq. (\ref{eq:3}), which stores the class-specific confusion for a class $c$ in the labeled pool $\mathbf{X}_{t}[k]$
\begin{align}
    \mathbf{Z}^c = \bigcup_{x \in \mathbf{X}_{t}[k]} \mathbf{P}^{c}\left(x,\mathbf{M}(\theta)\right)
    \label{eq:3}
\end{align}
Once we have $\mathbf{Z}^c$, we compute $\Lambda_{t}^{c}$, a $c^{th}$ class representative anchor, by computing average over $\mathbf{Z}^c$
%performing $k$-means clustering over $\mathbf{Z}^c$
\begin{align}
    \Lambda_{t}^{c} = average(\mathbf{Z}^c).
\end{align}
Each target class anchor $\Lambda_{t}^{c}$ serves as a representative of the $c^{th}$ class in the feature space. Now, in order to identify classes that are informative in the sense of class confusion in each unlabeled image $\mathcal{I}$
\begin{align}
    D_{(\mathcal{I})}^{c} =  \Vert\Lambda_{t}^{c} - \mathbf{P}^{c}(\mathcal{I},\mathbf{M}(\theta))\Vert_{2}; \mathcal{I}\in \mathbf{S}_{t}^{n_{b}}, c \in C
    \label{eq:5}
\end{align}
where, $\Vert \cdot \Vert_{2}$ denotes the $L_{2}$ norm of a vector. $D_{(\mathcal{I})}$ stores the disparity between the class anchors $\Lambda_{t}^{c}$ and class specific confusion for image $\mathcal{I}$. We say that a class $c$ in $\mathcal{I}$ is selected to be labeled if $D_{\mathcal{I}}^{c} > \delta$, where $\delta$ is a threshold set to $0.5$.

\begin{figure*}
	\centering
	\includegraphics[width=0.65\linewidth, height=3.5cm]{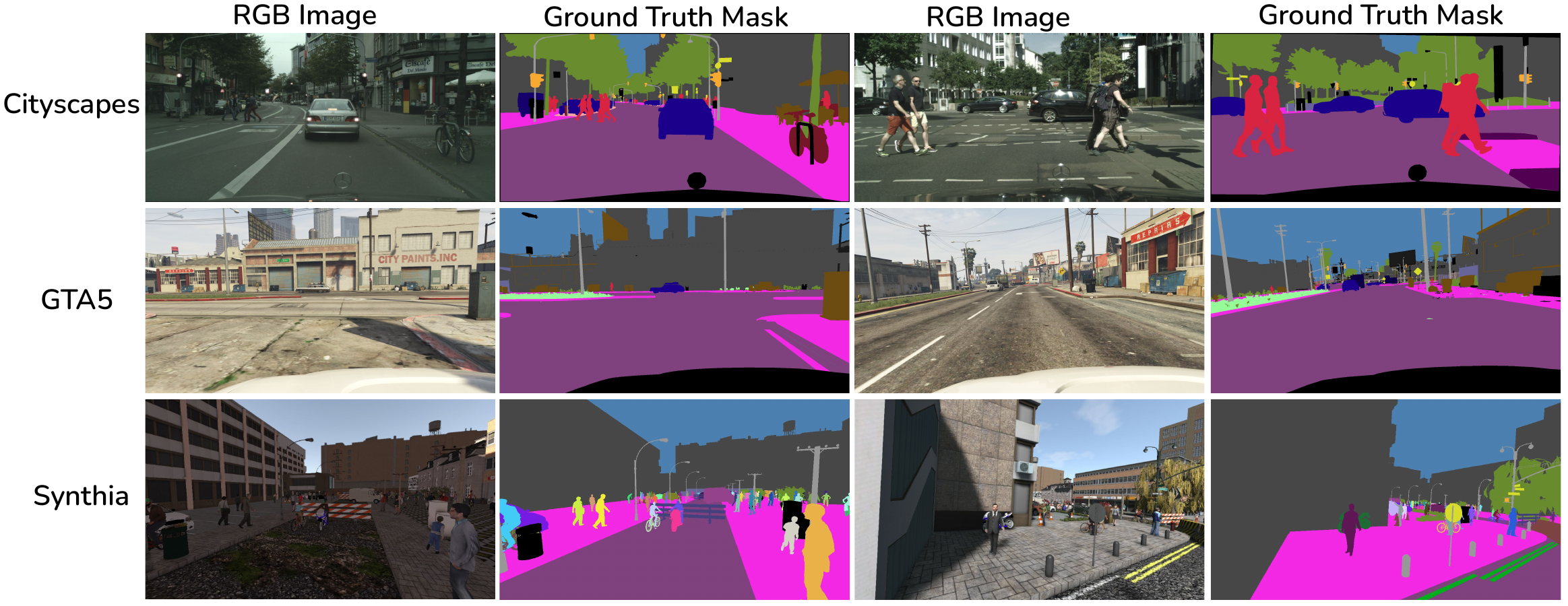}
	\caption{Representative RGB images with ground truth mask from the three datasets used: \cityscapes (first row) \gta (second row), \synthia (third row)}
	\label{fig:dataset-samples}
\end{figure*}
\begin{table*}[t]
	\centering
	\caption{Comparison with state-of-the-art DA techniques on GTA5$\rightarrow$Cityscapes. Number in bracket represents $\%$ of annoatated data.}
	\footnotesize
	\vspace{0.1cm}
	\begin{tabular}{m{2.6cm}|m{0.26cm}m{0.26cm}m{0.26cm}m{0.26cm}m{0.26cm}m{0.26cm}m{0.26cm}m{0.26cm}m{0.26cm}m{0.26cm}m{0.26cm}m{0.26cm}m{0.26cm}m{0.26cm}m{0.26cm}m{0.26cm}m{0.26cm}m{0.26cm}m{0.35cm}|m{0.35cm}}
	    \toprule[1pt]
	   \multicolumn{21}{c}{GTA5 $\rightarrow$ Cityscapes}\\\midrule
        Method & \rot{Road} & \rot{Sidewalk} & \rot{Building} & \rot{Wall} & \rot{Fence} & \rot{Pole} & \rot{T.light} & \rot{T.sign} & \rot{Vege} & \rot{Terrain} & \rot{Sky} & \rot{Person} & \rot{Rider} & \rot{Car} & \rot{Truck} & \rot{Bus} & \rot{Train} & \rot{Motorbike} & \rot{Bicycle} & mIoU \\\midrule
        %Source only &75.8 & 16.8 & 77.2 & 12.5 & 21   & 25.5 & 30.1 & 20.1 & 81.3 & 24.6 & 70.3 & 53.8 & 26.4 & 49.9 & 17.2 & 25.9 & 6.5  & 25.3 & 36   & 36.6 \\
        AdaptNet\cite{adaptnet} & 86.5 & 36   & 79.9 & 23.4 & 23.3 & 23.9 & 35.2 & 14.8 & 83.4 & 33.3 & 75.6 & 58.5 & 37.6 & 73.7 & 32.5 & 35.4 & 3.9  & 30.1 & 28.1 & 42.4 \\
        AdvEnt\cite{advent} & 89.4 & 33.1 & 81   & 26.6 & 26.8 & 27.2 & 33.5 & 24.7 & 83.9 & 34.7 & 78.8 & 58.7 & 30.5 & 84.8 & 38.5 & 44.5 & 1.7  & 31.6 & 32.4 & 45.5 \\
        CBST\cite{cbst} & 91.8 & 53.5 & 80.5 & 32.7 & 21.0   & 34.0   & 28.9 & 20.4 & 83.9 & 34.2 & 80.9 & 53.1 & 24.0  & 82.7 & 30.3 & 35.9 & 16.0  & 25.9 & 42.8 & 45.9 \\
        PyCDA\cite{pycda} & 90.5 & 36.3 & 84.4 & 32.4 & 28.7 & 34.6 & 36.4 & 31.5 & 86.8 & 37.9 & 78.5 & 62.3 & 21.5 & 85.6 & 27.9 & 34.8 & 18.0 & 22.9 & 49.3 & 47.4 \\
        IAST\cite{iast} & 94.1 & 58.8 & 85.4 & 39.7 & 29.2 & 25.1 & 43.1 & 34.2 & 84.8 & 34.6 & 88.7 & 62.7 & 30.3 & 87.6 & 42.3 & 50.3 & 24.7 & 35.2 & 40.2 & 52.2 \\
        WDA\cite{wda} & 94.0 & 62.7 & 86.3 & 36.5 & 32.8 & 38.4 & 44.9 & 51.0 & 86.1 & 43.4 & 87.7 & 66.4 & 36.5 & 87.9 & 44.1 & 58.8 & 23.2 & 35.6 & 55.9 & 56.4 \\
        ProDA\cite{proDA} & 87.8 & 56.0 & 79.7& 46.3 & 44.8 & 45.6 & 53.5 & 53.5 & 88.6 & 45.2 & 82.1 & 70.7 & 39.2 & 88.8 & 45.5 & 59.4 & 1.0 & 48.9 & 56.4 & 57.5\\
        %\textbf{Ours}        & 150    &      &      &      &      &      &      &      &      &      &      &      &      &      &      &      &      &      &      &      &     \\
        %DeepLabV2\cite{deeplabv2} & - & 96.5 & 77.8 & 88.9 & 49.1 & 51.2 & 52.4 & 53.1 & 63.8 & 90.6 & 57 & 91.8 & 72.9 & 40.8 & 92.1 & 64.3 & 65.3 & 49.5 & 44.0 & 68.2 & 66.8\\\hline
        CAG\cite{cag_uda} & 90.4 & 51.6 & 83.8 & 34.2 & 27.8 & 38.4 & 25.3 & 48.4 & 85.4 & 38.2 & 78.1 & 58.6 & 34.6 & 84.7 & 21.9 & 42.7 & 41.1 & 29.3 & 37.2 & 50.2 \\\hline
        AADA\cite{aada} (5\%)  & 92.2 & 59.9 & 87.3 & 36.4 & 45.7 & 46.1 & 50.6 & 59.5 & 88.3 & 44.0   & 90.2 & 59.7 & 38.2 & 90.0 & 55.3 & 45.1 & 32.0 & 32.6 & 52.9 & 49.3 \\
        MADA\cite{mada} (5\%) & 95.1 & 69.8 & 88.5 & 43.3 & 48.7 & 45.7 & 53.3 & 59.2 & 89.1 & 46.7 & 91.5 & \textbf{73.9} & \textbf{50.1} & 91.2 & 60.6 & 56.9 & 48.4 & \textbf{51.6} & \textbf{68.7} & 64.9 \\\hline
        %\rowcolor{lightgray}\textbf{Aug-based} (4.8\%)    & 96.9 & 76.4 & 89.1 & 46.8 & 48.5 & 47.0 & 54.9 & 60.9 & 90.0 & 52.7 & 92.7 & 71.1 & 38.6 & 91.9 & 62.2 & 60.0 & 53.8 & 47.0 & 67.3 & \textbf{65.5} \\
        \rowcolor{lightgray}\textbf{\texttt{Aug-based}} (4.8\%) & \textbf{97.5} & \textbf{76.9} & \textbf{90.2} & 46.5 & 48.9 & \textbf{47.9} & \textbf{55.9} & 61.2 & \textbf{90.3} & 52.9 & \textbf{93.2} & 71.6 & 40.6 & \textbf{92.1} & 62.9 & 60.9 & 52.6 & 47.5 & 67.9 &  \textbf{66.2} \\
        %\rowcolor{lightgray}\textbf{Anchor-based}(4.7\%)&96.9 & 77.1 & 89.3 & 48.5 & 47.4 & 47.4 & 54.5 & 58.2 & 89.4 & 52.7 & 93.2 & 71.1 & 41.3 & 91.4 & 70.1 & 64.6 & 46.8 & 48.3 & 66.1 & \textbf{66.1}    \\
        \rowcolor{lightgray}\textbf{\texttt{Anchor-based}}(4.7\%) & 96.9 & 76.6 & 88.8 & \textbf{47.9} & \textbf{49.0} & 47.1 & 55.2 & \textbf{61.6} & 89.8 & \textbf{55.3} & 92.9 & 70.9 & 39.3 & 92.0 & \textbf{63.3} & \textbf{63.7} & \textbf{58.6} & 48.7 & 66.7 & \textbf{66.6}    \\\hline
        \textbf{IoU-based}(4.5\%)& 96.9 & 78.1 & 89.4 & 47.6 & 50.0 & 49.1 & 55.8 & 61.4 & 89.9 & 54.4 & 93.6 & 70.9 & 47.0 & 91.6 & 70.6 &  67.9 & 49.6 & 47.8 & 67.2 & 67.4    \\
        Supervised & 97.3 & 80.6 & 90.1 & 53.2 & 54.8 & 51.0 & 55.4 & 64.0 & 90.5 & 55.1 & 93.3 & 74.3 & 51.0 & 92.7 & 75.7 & 76.5 & 55.2 & 42.9 & 68.0 & 69.6\\
        
	\bottomrule[1pt]
	\end{tabular}
	\label{tab:gta_city}
\end{table*}

\subsection{{\tt Aug-Based} Class Selection}
\label{sec:aug-based}
In the proposed augmentation based method, the core idea is to capture the disagreement in the class confusion from model predictions over strong and weakly augmented data. Strong augmentations are intended to make the predictions harder, which helps in identifying classes which are more difficult for the model to learn. We use strong and weak augmentations at test time with the same model $\mathbf{M}(\theta)$. For weak augmentations, we use transforms like \{\textit{hflip(0.5)}\}, while we use \{\textit{brightness (0.3), saturation (0.1), contrast(0.3), hflip (0.5), rotate (0.2)}\} for strong augmentations. 

For an image $\mathcal{I} \in \mathbf{S}_{t}^{n_{b}}$, we extract class confusion using eq. (\ref{eq:1}), and compute the disparity amongst class confusion from weak and strong augmented models as follows
\begin{align}
    D_{(\mathcal{I})}^{c} = \Vert\mathbf{P}^{c}(\mathcal{I}_w,\mathbf{M}(\theta)) - \mathbf{P}^{c}(\mathcal{I}_s,\mathbf{M}(\theta)\Vert_{2}; c \in C
\end{align}
where $\mathcal{I}_w$ and $\mathcal{I}_s$ denote the weakly and strongly augmented versions of the image $\mathcal{I}$, respectively. A class $c$ in $\mathcal{I}$ is selected to be labeled if $D_{\mathcal{I}}^{c} > \delta$, where $\delta$ is a threshold set to $0.5$.

\begin{algorithm}[t!]
    \caption{Algorithm for class selection}
    \label{algo:algorithm}
    \textbf{Given:} Segmentation Model $\mathbf{M}(\theta)$, unlabeled target domain $\mathcal{X}_{t}^{n_{t}}$, frame selection function $\Delta$, budget $n_{b}$, $\mathbf{X}_{t}[0] = \phi$, \\
    %\textbf{Output:} $D^{c}_{\mathcal{I}}$\\
    \textbf{Stage 1 (Active Labeling):}
    \begin{algorithmic}[1]
    \State Warmup $\mathbf{M}(\theta)$ with adversarial UDA \cite{adaptnet} weights
    \State $\mathbf{X}_{t}[1]^{n_{b}} = \Delta(\mathcal{X}_{t},n_{b})$ 
    \State Fine-tune $\mathbf{M}(\theta)$ using $\mathbf{X}_{t}[1]^{n_{b}}$
    \State for k = 2:n
    \State \quad\quad $\mathbf{S}_{t}^{n_{b}} = \Delta(\mathcal{X}_{t}^{n_{t}} \setminus \mathbf{X}_{t}[k-1])$; 
    \State \quad\quad Selecting class to be labeled for an image $\mathcal{I} \in \mathbf{S}_{t}^{n_{b}}$
    \State \quad\quad if class selection == \texttt{Anchor-based}:
    \State \quad\quad\quad\quad $\mathbf{Z}^{c} = \bigcup_{x\in \mathbf{X}_{t}[k]}\mathbf{P}^{c} (x,\mathbf{M}(\theta))$
    \State \quad\quad\quad\quad $\Lambda_{t}^{c} = average(\mathbf{Z}^{c}).$
    \State \quad\quad\quad\quad $D_{(\mathcal{I})}^{c} =  \parallel\Lambda_{t}^{c} - \mathbf{P}^{c}(\mathcal{I},\mathbf{M}(\theta))\parallel_{2}$; %\quad $\mathcal{I} \in \mathbf{S}_{t}^{n_{b}}$
    \State \quad\quad if class selection == \texttt{Aug-based}:
    \State \quad\quad\quad\quad $D_{(\mathcal{I})}^{c} = \parallel\mathbf{P}^{c}(\mathcal{I}_{w},\mathbf{M}(\theta)) - \mathbf{P}^{c}(\mathcal{I}_{s},\mathbf{M}(\theta)\parallel_{2}$
    \State \quad\quad if class selection == \texttt{IoU-based}:
    \State \quad\quad\quad\quad $D_{(\mathcal{I})}^{c} = (IoU_{\mathcal{I}}^{c}(\mathbf{M}(\theta)_{sup}) - IoU_{\mathcal{I}}^{c}(\mathbf{M}(\theta)))$
    \State \quad\quad A class $c$ in $\mathcal{I}$ is labeled if $D_{\mathcal{I}^{c}} > \delta(0.5)$
    \State \quad$\mathcal{L}_{CE} = \frac{1}{\mathcal{|I|}}\sum_{i\in \mathcal{I}}\sum_{c=1}^{C} \hat{Y}^{c} \log \mathbf{p}^{c}[i]$\\
    \textbf{Stage 2 (Self-training):}
    \State \quad$\mathcal{L}_{pseudo} = \mathcal{L}_{CE}(\{\mathcal{X}_{t}^{n_{t}}\setminus \mathbf{X}_{t}[k]\}, \hat{y}_{t})$
    \State \quad$\mathcal{L}_{seg} = \mathcal{L}_{CE}(\mathbf{X}_{t}[k],y_{t}) + \mathcal{L}_{pseudo}$
    \end{algorithmic}
    
\end{algorithm}

\subsection{{\tt IoU-Based} Class Selection}
\label{sec:iou-based}
We use \texttt{IoU-based} class selection as a setting to establish a skyline performance. Here, we assume that we have a model $M(\theta)_{sup}$ trained on the entire labeled target domain. The class selection is now based on the discrepancy between the class-wise IoU score between $M(\theta)$ (which is trained on $\mathbf{X}_t[k]$ in the $k^{th}$ AL iteration) and that of $M(\theta)_{sup}$. Bigger the gap between the two class-wise IoU scores, harder is the class.% for the current model. %The idea is to train the model with classes for which the model is not confident measured by its IoU score. So, given a model $\mathbf{M}(\theta)$ trained on initial labeled set $\mathbf{X}_{t}^{n_{b}}$ and $\mathbf{M}(\theta)_{sup}$ trained on entire labeled $\mathcal{X}_{t}^{n_{t}}$. 
For an image $\mathcal{I} \in \mathbf{S}_{t}^{n_{b}}$, we construct $D_{\mathcal{I}}^{c}$, measuring the difference in IoU score for a class $c$ when predicted using $\mathbf{M}(\theta)$ and $\mathbf{M}(\theta)_{sup}$
\begin{align}
    D_{(\mathcal{I})}^{c} = (IoU_{\mathcal{I}}^{c}(\mathbf{M}(\theta)_{sup}) - IoU_{\mathcal{I}}^{c}(\mathbf{M}(\theta)));c\in C
\end{align}
A class $c$ in $\mathcal{I}$ is selected to be labeled if $D_{\mathcal{I}}^{c} > \delta$, where $\delta$ is a threshold set to $0.5$.

\begin{table*}[t]
	\centering
	\footnotesize
	\caption{Comparison with state-of-the-art DA techniques on Synthia$\rightarrow$Cityscapes. Number in bracket represents $\%$ of annoatated data.}
	\vspace{0.1cm}
	\begin{tabular}{m{3cm} m{0.3cm} m{0.3cm} m{0.3cm} m{0.3cm} m{0.3cm} m{0.3cm} m{0.3cm} m{0.3cm} m{0.3cm} m{0.3cm} m{0.3cm}m{0.3cm}m{0.3cm}m{0.3cm}m{0.3cm}m{0.35cm} p{0.4cm}p{0.6cm}}
	    \toprule[1pt]
%	   \multicolumn{19}{c}{Synthia $\rightarrow$ Cityscapes}\\\midrule
        Method & \rot{Road} & \rot{Sidewalk} & \rot{Building} & \rot{Wall*} & \rot{Fence*} & \rot{Pole*} & \rot{T.light} & \rot{T.sign} & \rot{Vege} & \rot{Sky} & \rot{Person} & \rot{Rider} & \rot{Car} & \rot{Bus} &  \rot{Motorbike} & \rot{Bicycle} & mIoU & mIoU* \\\midrule
        %Source only & 64.3 & 21.3 & 73.1 & 2.4 & 1.1 & 31.4 & 7.0 & 27.7 & 63.1 & 67.6 & 42.2 & 19.9 & 73.1 & 15.3 & 10.5 & 38.9 & 34.9 & 40.3 \\ 
        AdaptNet\cite{adaptnet} & 79.2 & 37.2 & 78.8 & - & - & - & 9.9 & 10.5 & 78.2 & 80.5 & 53.5 & 19.6 & 67.0 & 29.5 & 21.6 & 31.3 & - & 45.9 \\
        AdvEnt\cite{advent} & 85.6 & 42.2 & 79.7 & 8.7 & 0.4 & 25.9 & 5.4 & 8.1 & 80.4 & 84.1 & 57.9 & 23.8 & 73.3 & 36.4 & 14.2 & 33.0 & 41.2 & 48.0 \\
        CBST\cite{cbst} & 68.0 & 29.9 & 76.3 & 10.8 & 1.4 & 33.9 & 22.8 & 29.5 & 77.6 & 78.3 & 60.6 & 28.3 & 81.6 & 23.5 & 18.8 & 39.8 & 42.6 & 48.9\\
        PyCDA\cite{pycda} & 75.5 & 30.9 & 83.3 & 20.8 & 0.7 & 32.7 & 27.3 & 33.5 & 84.7 & 85.0 & 64.1 & 25.4 & 85.0 & 45.2 & 21.2 & 32.0 & 46.7 & 53.3 \\
        IAST\cite{iast} & 81.9 & 41.5 & 83.3 & 17.7 & 4.6 & 32.3 & 30.9 & 28.8 & 83.4 & 85.0 & 65.5 & 30.8 & 86.5 & 38.2 & 33.1 & 52.7 & 49.8 & 57.0\\
        ProDA\cite{proDA} & 87.8 & 45.7 & 84.6 & 37.1 & 0.6 & 44.0 & 54.6 & 37.0 & 88.1 & 84.4 & 74.2 & 24.3 & 88.2 & 51.1 & 40.5 & 45.6 & 55.5 & 62.0 \\
        WDA\cite{wda} & 94.9 & 63.2 & 85.0 & 27.3 & 24.2 & 34.9 & 37.3 & 50.8 & 84.4 & 88.2 & 60.6 & 36.3 & 86.4 & 43.2 & 36.5 & 61.3 & 57.2 & 63.7 \\
        %DeepLabV2\cite{deeplabv2} & - & 97.4 & 79.5 & 90.3 & 51.1 & 52.4 &
        %49.0 & 57.5 & 68.0 & 90.5 & 93.1 & 75.1 & 53.9 & 92.7 & 80.2 & 58.1 & 71.1 & 72.5 & 77.5\\ \hline
        CAG\cite{cag_uda} & 84.7 & 40.8 & 81.7 & 7.8 & 0.0 & 35.1 & 13.3 & 22.7 & 84.5 & 77.6 & 64.2 & 27.8 & 80.9 & 19.7 & 22.7 & 48.3 & 44.5 & 50.9 \\\hline
        AADA\cite{aada} (5\%) & 91.3 & 57.6 & 86.9 & 37.6 & 48.3 & 45.0 & 50.4 & 58.5 & 88.2 & 90.3 & 69.4 & 37.9 & 89.9 & 44.5 & 32.8 & 62.5 & 61.9 & 66.2\\
        MADA\cite{mada} (5\%) & 96.5 & 74.6 & 88.8 & 45.9 & 43.8 & 46.7 & 52.4 & 60.5 & 89.7 & 92.2 & 74.1 & 51.2 & 90.9 & 60.3 & 52.4 & 69.4 & 68.1 & 73.3\\\hline
        \rowcolor{lightgray}\textbf{\texttt{Aug Based}} (4.9\%) & 97.1& 78.8 & \textbf{90.9} & \textbf{48.4} & 45.7 & \textbf{48.9} & 50.4 &65.5 & \textbf{90.7} & \textbf{93.2} & 75.9 & 49.9 & \textbf{92.7} & 69.9 & 52.9 & \textbf{71.4} &\textbf{70.1}&\textbf{75.3} \\
        \rowcolor{lightgray}\textbf{\texttt{Anchor Based}} (4.7\%) & \textbf{97.2}  & \textbf{79.6} & 90.5 & 45.5 & \textbf{50.8} & 48.7 & \textbf{55.4} & \textbf{67.1} & 90.2 & 93.2 & \textbf{76.1} & \textbf{53.2} & 90.1 & \textbf{73.3} & \textbf{53.9} & 70.4 &\textbf{70.9}&\textbf{76.1}  \\\hline
        \textbf{IoU Based} (5.1\%) & 97.7 & 80.9 & 90.8 & 49.1 & 56.1 & 52.3 & 59.1 & 68.5 & 90.8 & 93.1 & 75.8 & 54.1 & 93.1 & 78.4 & 56.7 & 71.1 &72.9 & 77.7  \\
        Fully Supervised & 97.6 & 81.3 & 91.1 & 49.8 & 57.6 & 53.8 & 59.6 & 69.1 & 91.2 & 94.4 & 76.7 & 55.6 & 93.3 & 79.9 & 57.7 & 72.2 & 73.8 & 78.4\\
	\bottomrule[1pt]
	\end{tabular}
	
	\label{tab:synthia_city}
\end{table*}

\subsection{Training Objective}
\label{sec:training}
Using all the actively labeled data, either by \texttt{Anchor-based} or \texttt{Aug-based} in the target domain, we can fine-tune the network to learn exclusive target domain information. Similar to \mada \cite{mada}, our training process comprises of two stages in each AL iteration. In the first stage, we use the standard cross entropy (CE) loss to train the network over the labeled data. 
% \begin{align}
% \mathcal{L}_{CE} = - \frac{1}{\mathcal{|I|}}\sum_{i\in \mathcal{I}}           \sum_{c=1}^{C} \hat{Y}^{c} \log \mathbf{p}^{c}[i]
% \end{align}
To further exploit the available unlabeled data, in the second stage we use a self-training approach using the \emph{pseudo-labels} obtained from the model trained in the first stage , such that $\hat{y}_{t} = \arg\max\mathbf{p}^{c}$ for the remaining unlabeled samples
\begin{align}
    \mathcal{L}_{pseudo} = \mathcal{L}_{CE}(\{\mathcal{X}_{t}^{n_{t}}\setminus \mathbf{X}_{t}[k]\}, \hat{y}_{t})
\end{align}
Thus the overall loss function for segmentation model is given as
\begin{align}
    \mathcal{L}_{seg} = \mathcal{L}_{CE}(\mathbf{X}_{t}[k],y_{t}) + \mathcal{L}_{pseudo} 
\end{align}
The overall training pipeline is summarized in Algo.\ref{algo:algorithm}.

\section{Dataset and Evaluation}

\mypara{Dataset}
For evaluation we use two common ``\textit{synthetic-2-real}'' segmentation setups as used in the contemporary \sota approaches \cite{mada, labor}, namely {\gtans$\rightarrow$\cityscapes} and {\synthians$\rightarrow$\cityscapes}. \gtans \cite {gta5} contains 24966 (1914x1052) images, sharing 19 classes with \cityscapes \cite{cityscapes}. \synthians \cite{synthia} contains 9400 (1280x760) images, sharing 16 classes. \cityscapes includes high resolution real world images of 2048x1024, with a split of 2975 training and 500 validation images. \cref{fig:dataset-samples} shows samples from the three datasets used to illustrate the significant distribution shift between the datasets.
%For further analysis we have used ``\textit{Real-2-Real}'' setup also. We adopted from Cityscapes$\rightarrow$BDD100k and Cityscapes$\rightarrow$IDD. BDD100k\cite{yu2020bdd100k} contains 7000 and 1000 in training and validation set, and IDD\cite{varma2019idd} an Indian road dataset consists of 6993 and 981 training and validation images respectively. 

\mypara{Implementation Details\footnote{Code:\url{https://github.com/sharat29ag/contextual_class}}}
We have followed the experimental setup of \mada \cite{mada}, and have used DeepLabV3+~\cite{deeplabv3+} with a ResNet-101 backbone for fair comparison. We have initialized the model with warm-up weights from AdaptNet~\cite{adaptnet}, an adversarial unsupervised domain adaptation framework. For training we have used 50 epochs with a batch size of 4 across all the experiments. For evaluation we have used \miou as a metric to measure model performance on \cityscapes validation set. We also report \emph{error margin} for various techniques defined as the difference between the particular ADA approach (at a certain annotation budget), and a fully supervised technique using the same backbone. 

\section{Experiments and Results}

%In this section we first demonstrate our results in comparison to \sota DA techniques, followed by the ablation experiments.
\begin{figure*}
    \centering
    \includegraphics[width=0.85\linewidth, height=4.5cm]{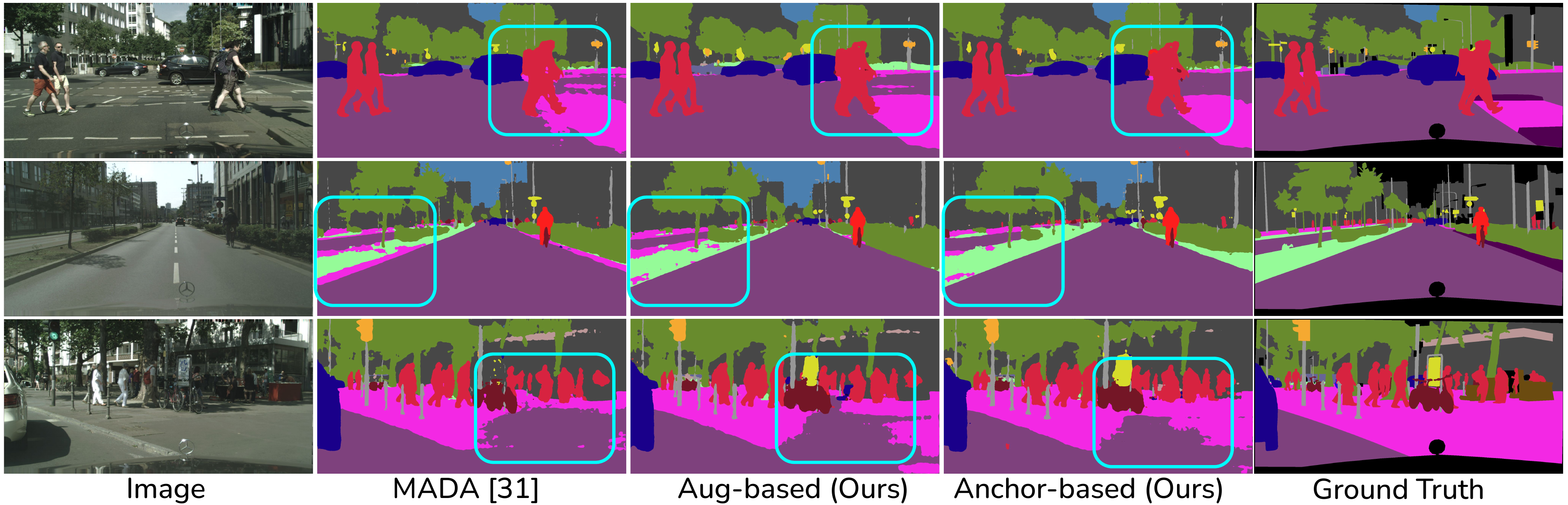}
    \caption{Qualitative results on \cityscapes Validation set after Domain Adaptation from {\gtans$\rightarrow$\cityscapes}. We compare our results of \texttt{Augmentation-based} and \texttt{Anchor-Based} with \mada\cite{mada}. We can clearly see the improvement in the highlighted regions of each image.}
    \label{fig:qualitative results}
\end{figure*}

\mypara{Quantitative Results}
We first show quantitative results on the two settings, {\gtans$\rightarrow$\cityscapes} and  {\synthians$\rightarrow$\cityscapes}, in Tables \ref{tab:gta_city} and \ref{tab:synthia_city} respectively. We compare our results with existing UDA~\cite{adaptnet,advent,cbst,proDA,iast}, semi-supervised~\cite{cag_uda}, weakly supervised~\cite{wda} and frame based ADA~\cite{aada,mada} techniques. We observe that both of our proposed approaches, \texttt{aug-based} and \texttt{anchor-based}, surpass the \sota techniques, reducing the error margin from $4.7$ to $3$ in {\gtans$\rightarrow$\cityscapes} and from $5.7$ to $3.2$ in {\synthians$\rightarrow$\cityscapes} using merely $5\%$ of the annotated data when compared to a fully supervised model.

\begin{table}[ht]
    \centering
    \footnotesize
    \caption{Ablation study on {\gtans$\rightarrow$\cityscapes} and {\synthians$\rightarrow$\cityscapes}, after training  using AL as well as pseudo-labels.}
    \begin{tabular}{lcc m{0.5cm} m{0.5cm}}
    \toprule[1pt]
    Method &Active Labels & Pseudo Labels & G$\rightarrow$C & S$\rightarrow$C   \\\midrule
    \rowcolor{lightgray}MADA \cite{mada} & $\checkmark$ & & 61.6 &65.0\\
    MADA \cite{mada} & $\checkmark$ & $\checkmark$ & 64.1 &68.1\\\hline
    \rowcolor{lightgray}\texttt{Aug-Based} &$\checkmark$ & & 65.5 & 75.1\\
    \texttt{Aug-Based} &$\checkmark$ &$\checkmark$  & 66.2 & 75.9\\ \hline
    \rowcolor{lightgray}\texttt{Anchor-Based} &$\checkmark$ & & 66.1 & 76.0\\ 
    \texttt{Anchor-Based} &$\checkmark$&$\checkmark$  & 66.6 & 76.5\\ \bottomrule[1pt]
    \end{tabular}
    \label{tab:ablation}
\end{table}

\mypara{Effectiveness of Proposed AL Strategies}
In \cref{tab:ablation} we break down the results of the two stages used in our pipeline: active learning, and pseudo labeling, and report \miou at each stage for both {\gtans$\rightarrow$\cityscapes} and {\synthians$\rightarrow$\cityscapes} setups. The purpose is to highlight the significant improvement of our active learning strategy over the \mada (row 1,3,5). In {\gtans$\rightarrow$\cityscapes} we observe an \miou improvement of $4.1$ ($61.6$ to $66.1$) and $3.9$ ($61.6$ to $65.5$) for \texttt{anchor-based} and \texttt{aug-based} approaches respectively. In future, we wish to work on improving our stage-2 performance by effectively using the pseudo labels complementary to our labeled samples. 

\mypara{Discussion}
\begin{enumerate*}[label=\textbf{(\arabic*)}]
	\item It is noteworthy that our proposed approach helps in increasing the \iou for most of the class labels. We also wish to highlight the simplicity of our approach in comparison to \mada \cite{mada} which selects target samples based on source anchors and multiple add-on components in the training process such as soft-anchor alignment loss, and updating target anchors with EMA. %In contrast, we use very few annotated sample along with pseudo-labels for the training, reducing the overall training time as well as increasing the accuracy.
	\item We observe reduction in annotation cost for both the proposed approaches, but \texttt{anchor-based} performs better. We speculate this is due to inability of \texttt{aug-based} approach in capturing contextual diversity. This is inline with the earlier works \cite{cdal} emphasizing the role  of contextual diversity in AL. 
\end{enumerate*}

\begin{figure*}[ht]
    \centering
    \includegraphics[width=0.9\linewidth, height=5.5cm]{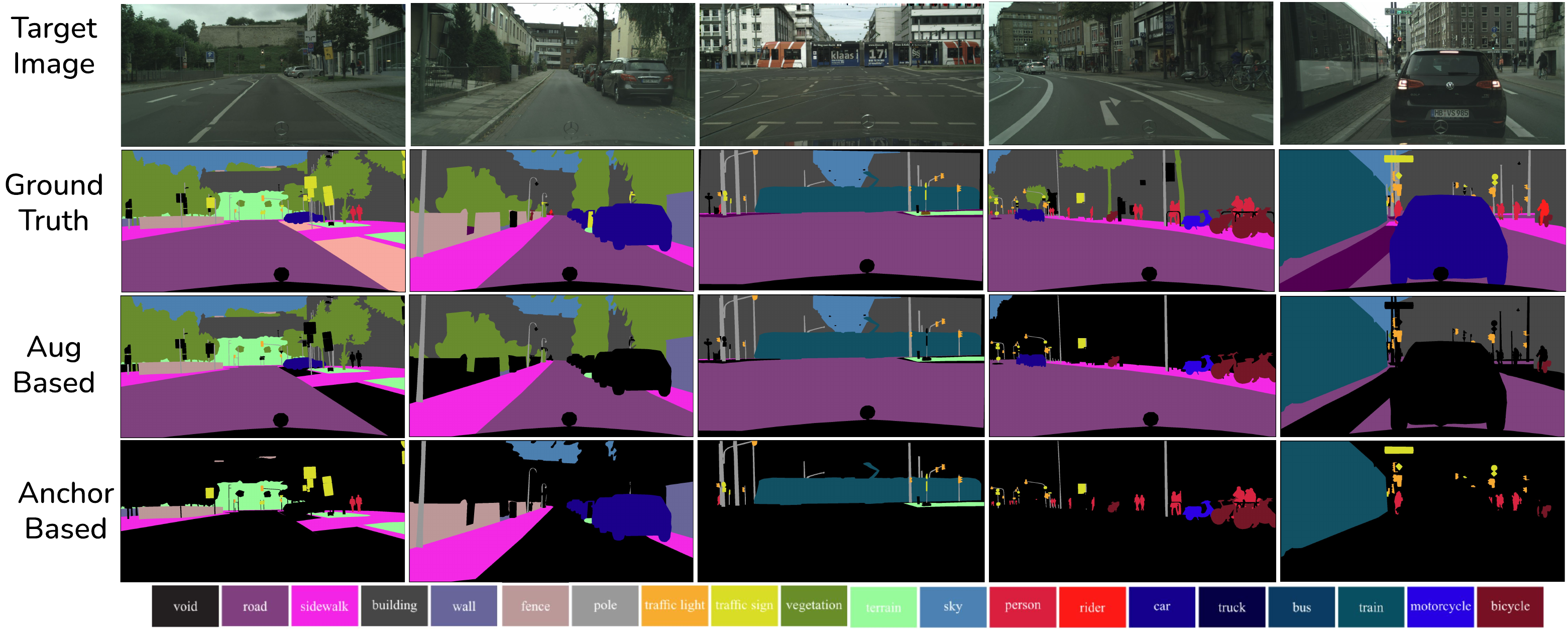}
    \caption{Qualitative samples of selecting classes using \texttt{Aug-based}(row-2) and \texttt{Anchor-based}(row3) approaches. Both the approaches reduces the annotation cost by selecting contextually diverse and informative classes. Notice, how the road regions frequently observed  in the target domain remains unlabeled in the frames selected by \texttt{Anchor-based} approach, but are labeled in \texttt{Aug-based} approaches due to the low confidence of the model.}%For example in col-3 and 5, Anchor based approach selects train and minor classes like traffic light, bicycle pole to strengthen their representation in the feature space, whereas Aug-based selects road and building also because of strong augmented model may not be confident in detecting it in the unseen rare context of train. Similarly in the crowded scenes like col-1 and 4, Aug-based techniques selects almost all the classes capturing the uncertainty, whereas Anchor-based reduces annotation cost by selecting important and semantically diverse classes. }
    \label{fig:qualitative_class_results}
\end{figure*}

\mypara{Result Visualization}
In \cref{fig:qualitative results} we visualize the output generated by our technique and compare it with the results of \mada. We can see in the highlighted regions that the predicted masks are more accurate and smooth in the confusing regions. 

\mypara{Selection Visualization}
In \cref{fig:qualitative_class_results} we show the selections made by two of our approaches (black color shows the non-selected regions). Notice, how the road regions frequently observed  in the target domain remains unlabeled in the frames selected by \texttt{Anchor-based} approach, but are labeled in \texttt{Aug-based} approaches due to the low confidence of the model because of appearance differences from the target. On the other hand, the label \texttt{fence} is less frequent in the target domain, and hence is selected by the \texttt{Anchor-based} approach. 

\begin{table}[t]
	\setlength{\tabcolsep}{5pt}
    \footnotesize
    \centering
    \caption{Results of \texttt{Aug-based}, and \texttt{Anchor-based} in conjunction with different frame selection techniques.}
    \begin{tabular}{m{1.4cm} m{0.8cm} m{0.6cm} m{0.8cm} m{0.6cm} m{0.8cm} m{0.6cm}}
    \toprule[1pt]
         \multirow{2}{*}{Method} & \multicolumn{2}{c}{Frame Based} & \multicolumn{2}{c}{\texttt{Aug-based}} & \multicolumn{2}{c}{\texttt{Anchor-based}} \\ 
         \cmidrule{2-3} \cmidrule{4-5} \cmidrule{6-7}
         & Data(\%) & \miou & Data(\%) & \miou & Data(\%) & \miou \\ \midrule
        Random & 3.4 & 57.8 & 2.8 & 60.2 & 2.1 & 60.5 \\
        Coreset \cite{coreset} & 3.4 & 58.8 & 2.9 & 60.5 & 2.4 & 61.2 \\
        MADA \cite{mada} & 3.4 & 60.3 & 2.6 & 62.5 & 2.3 & 62.9 \\
        CDAL \cite{cdal} & 3.4 & 62.04 & 2.8 & 63.8 & 2.2 & 63.5 \\
	\bottomrule[1pt]
    \end{tabular}
    \label{tab:ablation_selection}
\end{table}

\begin{figure}[ht]
    \centering
    \includegraphics[width=\linewidth, height=7cm]{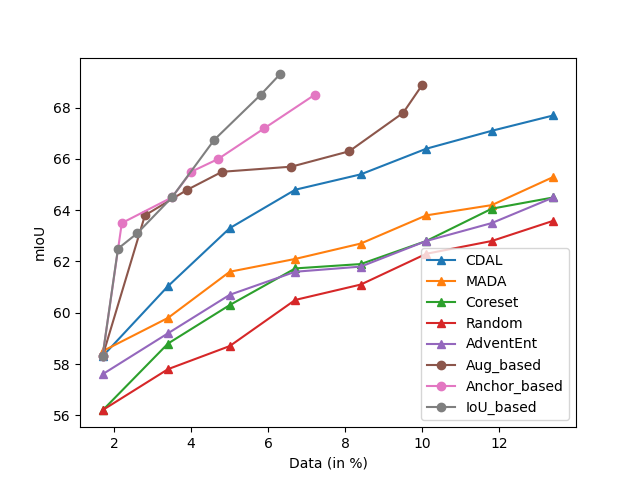}
    \caption{Comparison of state-of-the-art ADA, and  AL techniques at different annotation budgets. We use \texttt{IoU-based} as a skyline since it uses whole supervised information. We observe significant improvement in \miou using  our techniques for all annotation budgets.}
    \label{fig:AL_baselines}
\end{figure}

\mypara{Ablation Study for Effect of Frame Selection Strategy}
As stated in \cref{sec:training}, we have used \cdal\cite{cdal} as our base frame selection technique over which we reduce the annotation effort by selecting informative classes. To understand the impact of frame selection strategy, we replace \cdal\cite{cdal} with other frame selection techniques, such as \texttt{Core-Set}~\cite{coreset}, \mada \cite{mada} and \texttt{Random} selection. \cref{tab:ablation_selection} shows the results. We observe a significant improvement in performance using  both our approaches for each of the frame selection strategies.

\mypara{Improvements Obtained at Various Annotation Budgets}
We measure the impact of using various AL strategies at various annotation budget levels for the ADA problem on the GTA5$\rightarrow$Cityscapes experimental setup. The challenge for each technique is to reach the performance supervised model, $69.6$, with minimum annotation budget. For frame based techniques we increase the budget of 50 frames ($1.7\%$) at each AL step. Similarly, for our approaches we select 50 frames at each active cycle using \cdal and annotate classes either using one \texttt{anchor-based}, \texttt{aug-based} or \texttt{IoU-based} techniques. We note that it is difficult to control the exact annotation budget at each step in our approaches as we are annotating certain selected classes entirely. We now briefly discuss the baselines used in this experiment:
\begin{enumerate}[noitemsep]
    \item \texttt{Random-sampling:} For each active learning budgets, samples are randomly selected from the unlabeled pool.
    \item \texttt{Coreset~\cite{coreset}:} A subset selection approach, using K-center greedy algorithm for selecting diverse samples. 
    \item \texttt{AdvEnt \cite{advent}:} Samples were selected using the entropy maps of each samples predicted using \cite{advent} in the target domain.
    \item \texttt{CDAL~\cite{cdal}:} Selects contextually diverse samples exploiting the contextual information among the frames. 
    \item \texttt{MADA~\cite{mada}:} Selects samples complementary to the source anchors. 
\end{enumerate}
\cref{fig:AL_baselines} shows the results. Both of our proposed approaches, \texttt{aug-based} and \texttt{anchor-based}, surpass the baselines with a significant margin. We also note that we are very close to \texttt{IoU-based} selection at low annotation budgets. We also observe that using only $10\%$ annotation our \texttt{anchor-based} approach is able to achieve the performance of full supervised moodel.% This is contrast to \cdal~\cite{cdal} which requires $13.5\%$ annotation. %, thus reducing annotation cost of $3.5\%$ pixels or $100$ frames. 

\begin{table}[t]
	\setlength{\tabcolsep}{18pt}
	\footnotesize
    \caption{Results on the Source Free domain adaptation setting, where source data is unavailable (due to privacy or other such constraints), but we have a model trained on the source data. Note that existing ADA approaches like \mada \cite{mada} fail in this setting due to critical  dependency on the source data.}
    \centering
    \begin{tabular}{lcc}
    \toprule[1pt]
         Method & Data(\%) & \miou \\ \midrule
         URMA \cite{sivaprasad2021uncertainty} & - &45.1 \\ 
         SFDA \cite{liu2021source} & - &43.16 \\
         SRDA \cite{bateson2020source} & - &45.8 \\
         SFDASS \cite{kundu2021generalize} & -  &53.4 \\
         \rowcolor{lightgray}\texttt{Aug-based} & 3.3 & \textbf{60.9}\\
         \rowcolor{lightgray}\texttt{Anchor-Based} & 3.0 & \textbf{61.7} \\\bottomrule[1pt]
    \end{tabular}
    \label{tab:source_free}
\end{table}

\mypara{Source Free Domain Adaptation}
To show the extended utility of our approach beyond ADA, we also compare our \texttt{Anchor-based} and \texttt{Aug-based} approaches in a Source-Free Domain Adaptation (\texttt{SFDA}) setting. In \texttt{SFDA}, the source dataset is unavailable due to privacy issues, but we have a segmentation model trained on the source dataset. The existing ADA technique, \mada\cite{mada} fails to adapt in the \texttt{SFDA} setting due to its dependency on the source data for computing source anchors to select the samples from target dataset. \cref{tab:source_free} shows the results comparing  with state-of-the-art SFDA approaches like \cite{sivaprasad2021uncertainty, liu2021source, bateson2020source, kundu2021generalize}. We observe an improvement of $7.5$ and $8.3$ using our two approaches, over the baseline of $53.4$, using only $3.3\%$ and $3.0\%$ of the annotated data.

\section{Discussion}

\myfirstpara{Conclusion}
In this paper, we have proposed a novel and intuitive approach to reduce annotation effort by labeling certain informative classes in a frame instead of wasting the annotation budget by labeling redundant regions. Through extensive experiments and comparison with different ADA and Active Learning baselines, we highlight the improvement in performance using both proposed \texttt{Aug-based} and \texttt{Anchor-based} approaches. We also validate that our approaches can be used as a decorator for any frame-based active learning approaches, which helps reduce annotation cost and increases the model performance beyond the existing state-of-the-art. 

\mypara{Societal Impact}
In today's era, collecting data is relatively easier than labeling, which often requires expensive and scarce expert guidance. This is even more important in other domains with more direct societal impact such as medical imaging, and e-governance. We hope to validate our work to these domains in future. 
%for Progress in fields like medical imaging and autonomous driving is restricted owing to the lack of labeled data. We feel that effectively proposing better solutions to train our networks using minimal but informative data is essential, which requires data repair or intelligent labeling. Thus we hope that further progress is made in reducing annotation costs and increasing the model performance by effectively utilizing the data. 

\mypara{Limitations}
While our work can effectively utilize a given annotation budget by selecting most informative samples, we have used some off-the-shelf techniques for generating  pseudo-labels. 
%In our work, we proposed two active labeling strategies, but we lack to utilize the unlabeled data effectively in a self-training manner. 
In future, we would like to explore generating pseudo labels complementary to the labeled data  from the target domain, as well as exploit label distribution and extracting useful information from the abundantly available source labeled data.

{\small
\bibliographystyle{ieee_fullname}
\bibliography{egbib}
}

\end{document}